\documentclass[runningheads]{llncs}
\usepackage{hyperref} 
\hypersetup{
    colorlinks=true,
    linkcolor=blue,
    filecolor=magenta,      
    urlcolor=cyan,
    pdftitle={Overleaf Example},
    pdfpagemode=FullScreen,
    }
\usepackage{cite}
\usepackage{hyperref}
\usepackage{ragged2e}
\usepackage[table]{xcolor}
\usepackage{amsmath,amssymb,amsfonts}
\usepackage{algorithmic}
\usepackage{graphicx}
\usepackage{fancyhdr}
\usepackage{caption}
\usepackage{subcaption}
\usepackage{textcomp}
\usepackage{tikz, pgfplots}
\pgfplotsset{compat=1.15}
\usepackage{xcolor}
\def\BibTeX{{\rm B\kern-.05em{\sc i\kern-.025em b}\kern-.08em
    T\kern-.1667em\lower.7ex\hbox{E}\kern-.125emX}}
\begin{document}
\title{Deep Learning Models for UAV-Assisted Bridge Inspection: A YOLO Benchmark Analysis}
%
%
\author{Trong-Nhan Phan\inst{1, 2} \and
Hoang-Hai Nguyen\inst{1, 2} \and
Thi-Thu-Hien Ha\inst{1, 2} \and
Huy-Tan Thai\inst{1, 2} \and
Kim-Hung Le\inst{1, 2}
}
\authorrunning{N.T. Phan et al.}
%
\institute{University of Information Technology, Ho Chi Minh City, Vietnam \and Vietnam National University, Ho Chi Minh City, Vietnam\\
\email{\{21522407, 21522034, 21522056\}@gm.uit.edu.vn, \{tanth, hunglk\}@uit.edu.vn}}
\maketitle              
\begin{abstract}
\justifying
Visual inspections of bridges are critical to ensure their safety and identify potential failures early. This inspection process can be rapidly and accurately automated by using unmanned aerial vehicles (UAVs) integrated with deep learning models. However, choosing an appropriate model that is lightweight enough to integrate into the UAV and fulfills the strict requirements for inference time and accuracy is challenging. Therefore, our work contributes to the advancement of this model selection process by conducting a benchmark of 23 models belonging to the four newest YOLO variants (YOLOv5, YOLOv6, YOLOv7, YOLOv8) on COCO-Bridge-2021+, a dataset for bridge details detection. Through comprehensive benchmarking, we identify YOLOv8n, YOLOv7tiny, YOLOv6m, and YOLOv6m6 as the models offering an optimal balance between accuracy and processing speed, with mAP@50 scores of 0.803, 0.837, 0.853, and 0.872, and inference times of 5.3ms, 7.5ms, 14.06ms, and 39.33ms, respectively. Our findings accelerate the model selection process for UAVs, enabling more efficient and reliable bridge inspections.

\end{abstract}
%
%
%


\section{Introduction} 

Visual inspection, a method using one’s naked eye to assess the condition of structures, has played an essential role in maintaining and ensuring the quality of urban public works~\cite{visual_inspection}. This method involves detecting structural issues such as corrosion, cracks, and other problems to ensure the works’ safety and durability. For instance, visual inspection is routinely used to inspect the condition of high-rise buildings’ envelopes. For bridges constantly subjected to harsh weather conditions and traffic loads, visual inspection is an excellent method for inspectors to detect early signs of deterioration, corrosion, and damage. As a result, experts can suggest timely solutions to maintain and repair these structures. 
Technology devices, like cameras, digital image analyzers, and other devices, have contributed to improving visual inspection. These tools allow experts to remotely examine components for corrosion and damage, which reduces cost and safety risks for inspectors in visual inspection~\cite{remote_vi}. Among them, UAVs are commonly used to inspect structures that are difficult to access or have complex geometries. However, the monitoring process by UAVs encounters particular challenges. First, during the visual inspection process, a large amount of data requires specific expertise to analyze and examine, which is labor intensive~\cite{challenge_vi_2}. Secondly, inspectors might face low-quality images due to the lack of proper light conditions and wind turbulent disturbance, leading to false detection~\cite{challenge_vi}. Therefore, it is necessary to automate the process with AI-supported to increase accuracy while reducing the inspection time.

The mentioned automatic process can be achieved using computer vision models. Computer vision (CV) is a branch of artificial intelligence that enables computers and systems to extract meaningful insights from digital images, videos, and other visual data sources. Subsequently, they can provide recommendations based on the information they gather. In visual inspection, automation can be done by combining a computer vision model with a camera-attached UAV. This could be a game-changer in the field of visual construction inspection as UAVs have adaptability, easy installation, low maintenance costs, versatility, and relatively small operating cost~\cite{Hentati2020}, while computer vision models are robust, easy to train and deploy. In this process, UAVs take pictures of structures. After that, the computer vision model processes the images and gives predictions about what components appear in the images and their possible problems. Then, this information would be sent to the inspectors, which would assist them in assessing the structure's condition and accordingly decide whether the structures need to be repaired or renovated. Currently, deep learning models are also the main topic of this paper.

Although the combination of computer vision models and camera-attached UAV(s) is the future of automated visual inspections, specific problems need to be examined. Despite the rapid advancement of UAVs in recent years, their computational power and storage space are still limited. Meanwhile, in visual inspection, the number of images that need to be processed could grow enormously, and data collection is often repeated since external factors, such as wind-induced vibrations in captured footage or low illumination leading to dark images, can significantly affect the quality of data \cite{Zakaria2022}. The size of images is also a significant problem since processing, storing, or transmitting such a large amount of data is time-consuming and storage-intensive. In the traditional centralized CV-UAV method, these problems are not properly resolved because CV models are placed on a local computer or in the cloud, and UAVs have an additional task to transmit these images to where CV models are placed and wait to receive the results. Having to transmit images between UAVs and CV models is a bottleneck of the process, as it is slow and vulnerable to poor connections and hackers’ attacks. Edge AI is a rising technology that can deal with that. It overcomes this problem by embedding CV models into UAVs, so images are processed inside the UAVs, and only the results are sent to inspectors. The results are often text and damaged-predicted images, just a tiny part of the captured images. By doing so, the amount of data needed to be transmitted is reduced by a lot, thus speeding up the process and minimizing the risk caused by poor connections and hackers.

We found that the task of discovering the potential of Edge AI in the field of visual inspection, especially of bridges, is essential~\cite{le2023brainyedge}. The principal reason is that bridges are sensitive infrastructure transportation components, so they must be routinely inspected. It is necessary to develop an automatic bridge inspection method combining Edge AI and visual inspections to quickly and correctly assess the condition of bridges while requiring low effort. This method would help the bridge preservation process go smoothly and hence ensure the safety of traffic participants as well as social connectivity. However, incorporating CV models into UAVs is a challenge. This is because some modern CV models achieve high accuracy but require enormous computational and storage capacity, which UAVs lack.
On the other hand, some older CV models fit UAVs' computational and storage capacity, but their accuracy is inadequate. Through this work, we aim to find suitable CV models for bridge inspection with UAVs and provide a benchmark for future works. 
Therefore, in this paper, we focus on experimenting with CV models, comparing their performance through accuracy and inference time on \textit{COCO-Bridge 2021+}, a dataset for bridge inspection. In detail, we benchmark four YOLO variants, including YOLOv5~\cite{Jocher_YOLOv5_by_Ultralytics_2020}, YOLOv6~\cite{https://doi.org/10.48550/arxiv.2209.02976}, YOLOv7~\cite{https://doi.org/10.48550/arxiv.2207.02696}, and YOLOv8~\cite{Jocher_Ultralytics_YOLO_2023}. The experimental results show that YOLOv8n, YOLOv7tiny, YOLOv6m, and YOLOv6m6 are the models that achieve the best trade-off between accuracy and inference time among 23 total models of these four YOLO variants.


\section{Related work} 
\label{section1}


The application of machine learning has been gaining significant attention in various fields, including agriculture, visual inspection within the petroleum industry, urban building maintenance, and bridge inspection. When it comes to the first field, agriculture, visual inspection plays an important role in identifying leaf diseases. In an article from 2021 \cite{Thai2021}, authors used a Vision Transformer model (ViT) integrated with a Drone Pi to categorize and group diseased leaves. The Cassava Leaf Disease Dataset used in this study includes $21,397$ photos of cassava leaves in five distinct groups: CBB, CBSD, CGM, CMD (four types of leaf diseases), and healthy leaves. The ViT model produced remarkable results and demonstrated its potential for early identification of leaf disease~\cite{thai2023formerleaf}. In detail, this model observed the highest F1 score of $96$\% in the detection of CMD. Another work \cite{Zhang2019} (2019) utilized a deep convolutional neural network (DCNN) to analyze hyperspectral images, taken by UAVs, with high spatial resolution. Specifically, in a well-controlled field experiment, drones gathered pictures on five different days on wheat fields that were either healthy or infected with rust across a whole crop cycle. Then, the model used these photos for calibration and got a high performance throughout the lifespan of plants, especially at the late stages of disease spreading. The proposed model obtained an accuracy, at $85$\%, better than the conventional random forest classification's performance, at $77$\%. The two aforementioned works contribute significantly to the advancement of the automation of crop disease detection.


In the area of building inspection, it is clear that using UAVs (Unmanned Aerial Vehicles) would be both cost-effective and safe. A 2021 study \cite{Ko2021} presented an automated inspection system for exterior crack detection on high-rise buildings using UAVs. A deep learning image classification algorithm based on Google's Xception CNN yielded an excellent performance in detecting wall cracks, with a Success Percentage at $80$\%, where this measure is calculated by the ratio between the number of cracks detected and the total number of cracks. The dataset has 20,001 non-crack images and 20,001 crack images, where each image is $227\times227$ pixels. Another study \cite{Katsigiannis2023}, in 2023, investigated using deep learning, with a small amount of data and transfer learning, to detect cracks in masonry facades. This approach achieved remarkable accuracy and F1 scores, reaching $100$\% using CNN models, specifically MobileNetV2 and InceptionResNetV2, on the Brickwork Cracks Dataset containing $5,926$ images.
In 2023, the authors in~\cite{Wang2023} employed three different deep learning models, ResNet-50, VGG-16, and AlexNet, to detect surface defects and reconstruct the 3D scene.
These mentioned studies show the potential to enhance the inspection efficiency in buildings using deep learning models. 
 
Finally, in bridge inspection, visual inspection is a pivotal element. Notably, in a 2018 study by the American Mechanical Engineers \cite{Zhao2018}, a method employing deep learning was suggested for comprehensive maintenance and inspection of bridges based on a dataset containing $3,832$ images collected from the internet, including arch, suspension, and cable-stayed bridges. This method utilized a fine-tuned AlexNet to classify bridge types, achieving validation and testing accuracies of $96$\% and $96.6$\%, respectively. In addition, a Faster R-CNN model based on ZF-net was utilized to identify and locate bridge towers and decks and obtained an mAP of $90.45$\% on the validation dataset. Detecting concrete cracks in bridges was achieved through a fine-tuned GoogLeNet, attaining high validation and testing accuracies of $99.39$\% and $99.36$\%, successively. In 2017, another noteworthy study \cite{Cha2017} presented a region-based deep learning strategy for autonomous structural visual inspection. Authors used a Faster R-CNN model to detect different types of damage in real-time, including bolt corrosion, concrete cracks, steel corrosion, and steel delamination on a dataset containing $2,366$ images labeled. The result showed impressive average accuracy (AP) ratings for each category, with an overall mean AP of $87.8$\%. Additionally, a 2019 article \cite{Zhang2019_a} introduced a single-stage detector for concrete bridge surface damage detection. The $2,206$ inspection photos of Hong Kong's highway concrete bridges were included in the image dataset used for this method. The original YOLOv3 model was enhanced through a novel transfer learning technique, significantly improving detection accuracy to $80$\% and $47$\% at the IoU threshold of $0.5$ and $0.75$, correspondingly. Using deep learning and texture analysis, ASR (Abstract Alkali-Silica Reaction) crack diagnosis is a new advancement in bridge inspection \cite{Nguyen2023}. A dataset containing $35$ ASR defect photos from different Queenslandian bridges impacted by ASR was produced for this investigation. The integration of texture morphology and the InceptionV3 model led to a remarkable validation accuracy of $94.07$\%. These results demonstrate the feasibility of machine learning for bridge inspection, especially damage detection.

In conclusion, our comprehensive review of research underscores the impactful role of machine learning and deep learning methods in advancing visual inspection processes across various industries. These state-of-the-art technologies not only excel in early identifying potential issues but also significantly enhance safety, productivity, and overall infrastructure quality.

\section{Models and Evaluation methodology}

This section serves as the cornerstone of our study, outlining the models and evaluation methodology. 
Regarding models, we introduce the four variant models of YOLO from version 5 to version 8 in Section~\ref{methoda}. For evaluation methodology, we present the experimental dataset, followed by implementation tools and model selection criteria, and end up with evaluation metrics in Section~\ref{methodb}.

\subsection{Deep learning models} 
\label{methoda}

Nowadays, two types of object detectors may be distinguished: one-stage detectors and two-stage detectors. In comparison to one-stage detectors, two-stage detectors frequently reach higher accuracy (both recognition and localization accuracy) and have slower inference times. Our work focuses entirely on one-stage detectors since they are suitable for real-time object detection.
Models tested in our work include the four newest: YOLOv5, YOLOv6, YOLOv7 and YOLOv8. Apart from the fact that they are state-of-the-art models for Object Detection, all of them have controversial performances, with opposite results in different scenarios. Thus, those models’ efficiency in the visual inspection of bridges is questionable and needs to be studied thoroughly.
\begin{itemize}
    
    \item \textbf{YOLOv5:} YOLOv5 was released in June 2020 by Glenn Jocher \cite{Jocher_YOLOv5_by_Ultralytics_2020}. YOLOv5 built upon many of the improvements made in YOLOv4, including flexible control of model size, application of Hardswish activation function, and data enhancement \cite{Jiang2022}, but the major difference is that it was built based on PyTorch framework instead of DarkNet. From the literature, we found that YOLOv5 is better than most SOTA on various datasets \cite{https://doi.org/10.48550/arxiv.2107.08430, Nepal2022, Liu2021}. However, some works show contradictory results. For instance, YOLOv4 yields better AP50 than YOLOv5 on a custom faulty components detection of electric poles dataset \cite{Rahman2021}. \\
    
    \item \textbf{YOLOv6:} YOLOv6 was released in 2022 by Meituan Vision AI Department \cite{https://doi.org/10.48550/arxiv.2209.02976}. It has many improvements in backbone, neck, head, and training strategies, with the most noticeable being its anchor-free paradigm. Previous works show that, compared to YOLOv5, YOLOv6 lacks stability and flexibility but makes up for impressive capabilities in detecting small, densely packed objects. Official YOLOv6 papers show that YOLOv6 outperforms YOLOv5 and all previous SOTA in almost all categories \cite{https://doi.org/10.48550/arxiv.2209.02976, https://doi.org/10.48550/arxiv.2301.05586}. Meanwhile, on a sampled COCO train2017 dataset, YOLOv6 has a higher recall and mAP@50:95 while having worse precision, mAP@50, and inference speed compared to YOLOv5 \cite{inproceedings}. \\ 

    \item \textbf{YOLOv7:} YOLOv7 was released in 2022 by Alexey Bochkovskiy et al. \cite{https://doi.org/10.48550/arxiv.2207.02696}. YOLOv7’s architecture has three main features:  E-ELAN for efficient learning, model scaling for varying sizes, and the "bag-of-freebies" approach for accuracy and efficiency. In fact, YOLOv7 is promising as many works show that it is better than all previous SOTA, especially in terms of accuracy \cite{inproceedingsyolov7, https://doi.org/10.48550/arxiv.2207.02696, Shokri2023, xinming2023comparative, Olorunshola2023, Nguyen2022}. 

    \item \textbf{YOLOv8:} YOLOv8 was released in 2023 by Ultralytics, the company behind YOLOv5 \cite{Jocher_Ultralytics_YOLO_2023}. YOLOv8 is anchor-free, predicts fewer boxes, and has a faster non-maximum suppression (NMS) process. As YOLOv8 was introduced recently, there are not many works studying it in the literature. Overall, YOLOv8 seems better than YOLOv5 (except for training speed) and has similar performance compared to YOLOv7 \cite{yolov578}. \\


\end{itemize}

\subsection{Evaluation methodology}
\label{methodb}

\subsubsection{Evaluation dataset}
\label{methodb2}

\begin{figure*}
    \centering
    \begin{subfigure}[b]{0.24\textwidth}
         \centering
         \includegraphics[width=\textwidth]{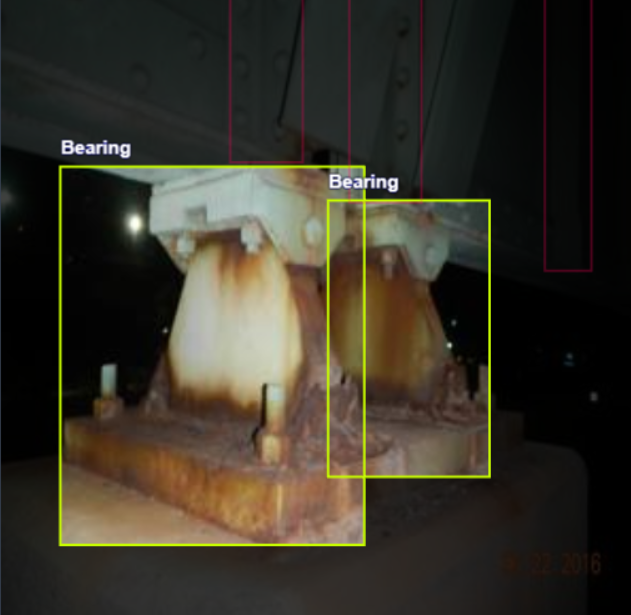}
         \caption{}
         \label{fig:bear}
     \end{subfigure}
     \hfill
     \begin{subfigure}[b]{0.24\textwidth}
         \centering
         \includegraphics[width=\textwidth]{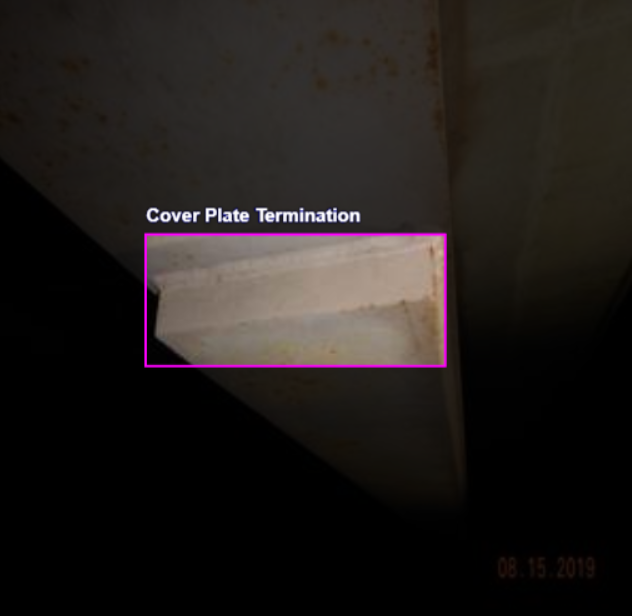}
         \caption{}
         \label{fig:cover}
     \end{subfigure}
     \hfill
     \begin{subfigure}[b]{0.24\textwidth}
         \centering
         \includegraphics[width=\textwidth]{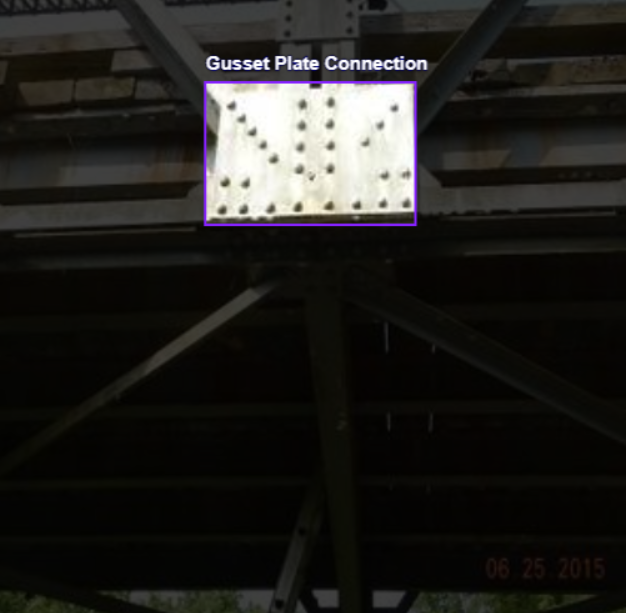}
         \caption{}
         \label{fig:gusset}
     \end{subfigure}
     \begin{subfigure}[b]{0.24\textwidth}
         \centering
         \includegraphics[width=\textwidth]{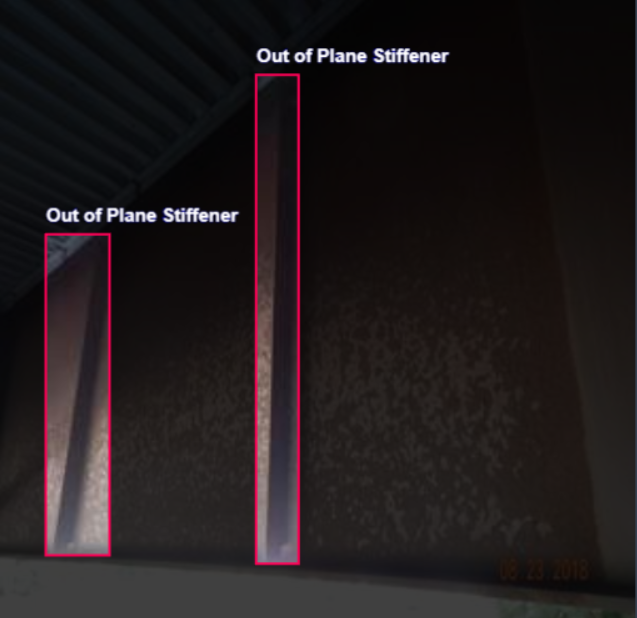}
         \caption{}
         \label{fig:outof}
     \end{subfigure}
    \caption{Sample images of each structural bridge detail: (a) bearing; (b) cover plate terminations;  (c) gusset plate connections; (d) out-of-plane stiffeners.}
    \label{fig:classes}
\end{figure*}

The evaluation dataset used in this paper is \href{https://data.lib.vt.edu/articles/dataset/COCO-Bridge_2021_Dataset/16624495/1}{COCO-Bridge 2021+ Dataset} authored by Eric Bianchi and Matt Hebdon. The Virginia Department of Transportation (VDOT) provided real structural bridge inspection records, which served as the origin of the dataset used for bounding box detection. In specifics, this dataset is made up of 1470 annotated structural bridge photos, of which about 10\% are from the test set (136 images) and 90\% are from the training set (1321 images). The dataset contains information on four structural bridge details: out-of-plane stiffeners, cover plate terminations, gusset plate connections, and bearings. The dataset's sample photos for each category are displayed in \autoref{fig:classes}.


\subsubsection{Implementations tools}
\label{methodb1}

All experiments were carried out using an identical computer with an x86\_64 CPU architecture, CUDNN version 9.2.1, and two CUDA devices. The GPU used was an NVIDIA GeForce RTX 3060 Ti with approximately 8.36 GB of memory. The software versions used were Torch version 2.0.1+cu117, Torchvision version 0.15.2+cu117, and Python version 3.9.18. 

\subsubsection{Evaluation Metrics}
\label{methodb4}

In this section, we introduce the purpose of evaluating the performance of our model in detecting the details of the bridge. It is important to note that accurate evaluation is crucial to ensure the model works effectively in real-world applications.

\begin{itemize}
\item \textbf{Intersection over Union (IoU)}: This measures the overlap between the predicted bounding box and the ground truth bounding box. IoU can be formulated as follows:
\begin{equation}
IoU = \frac{Area\_of\_Overlap}{Area\_of\_Union}
\end{equation}

\item \textbf{Average Precision (AP)}: This summary of the shape of the Precision-Recall curve to a single number simplifies the model evaluation task. All models in our work use Object-Detection-Metrics's Every Points Interpolation AP. 

\item \textbf{Mean Average Precision (mAP)}: This is a popular metric to measure the performance of object detection models. It calculates the average of average precision (AP) over all classes. There are two mAP metrics used in our work: mAP@50 and mAP@50:95. mAP@50 calculates the mAP at an IoU threshold of $0.5$, while mAP@50:95 consider different IoU thresholds from $0.5$ to $0.9$. 

\item \textbf{Inference Time}: Refers to the duration it takes a model to process an input and generate a corresponding output. This encompasses the entire process, from when the model receives the input to when it delivers the final result. This metric is crucial for real-time applications and is influenced by both the model's structure and the available computational resources.

\item \textbf{Number of parameters}: The number of parameters can affect both the computational efficiency and the capacity of the model. A model with too many parameters may overfit the training data, while a model with too few parameters may not be able to learn the underlying patterns in the data.
 
\item \textbf{GFLOPs}: This stands for Giga Floating Point Operations, and it measures the computational complexity of the model. It is the total number of floating point operations (additions, multiplications, etc.) required to perform a forward pass through the model. A model with fewer GFLOPs is generally faster and consumes less energy. \end{itemize}

\section{Experimental Results}
\label{results}



 In this section, we describe our experimental results on the COCO-bridge-2021+ dataset. \autoref{tab:result1} illustrates the input size, the number of parameters, and GFLOPs of YOLO variants, indicating that models are available in a broad range of input sizes and complexity. In detail, models take 640x640 pixels or 1280x1280 pixels input sizes and have parameter numbers in the approximate range of 1.76M - 152.90M and GFLOPs in the range of 4.1 - 672.5. The input size of 640x640 pixels is enough for the models to detect most objects in an image, but the larger input size of 1280x1280 pixels may help in the case of blurred images or far-away objects at the cost of higher model complexity.
 
\begin{table}[tb!]
    \centering
    \caption{The summary of models's input size, param, and GFLOPs.}
    \label{tab:result1}
    
    \rowcolors{2}{gray!25}{white}
    \begin{tabular}{|c|c|c|c|} \hline 
        \rowcolor{gray!50}
        \textbf{Model} & \textbf{Input size (pixels)} & \textbf{Params (M)} & \textbf{GFLOPs} \\
        \hline
        YOLOv5n & 640 x 640 & 1.76 & 4.1\\
        YOLOv5s & 640 x 640 & 7.02 & 15.8\\
        YOLOv5m & 640 x 640 & 20.87 & 47.9\\
        YOLOv5l & 640 x 640 & 46.12 & 107.7\\
        YOLOv5x & 640 x 640 & 86.19 & 203.8\\
        YOLOv6n & 640 x 640 & 4.63 & 11.3\\
        YOLOv6s & 640 x 640 & 18.50 & 45.2\\
        YOLOv6m & 640 x 640 & 34.80 & 85.6\\
        YOLOv6l & 640 x 640 & 59.54 & 150.5\\
        YOLOv6n6 & 1280 x 1280 & 10.34 & 12.4\\
        YOLOv6s6 & 1280 x 1280 & 41.32 & 197.5\\
        YOLOv6m6 & 1280 x 1280 & 79.53 & 378.8\\
        YOLOv6l6 & 1280 x 1280 & 140.21 & 672.5\\
        YOLOv7tiny & 640 x 640 & 6.02 & 13.0\\
        YOLOv7 & 640 x 640 & 36.50 & 103.2\\
        YOLOv7x & 640 x 640 & 70.80 & 188.0\\
        YOLOv7w6 & 1280 x 1280 & 69.81 & 88.6\\
        YOLOv7e6 & 1280 x 1280 & 110.39 & 143.6\\
        YOLOv7d6 & 1280 x 1280 & 152.90 & 197.4\\
        YOLOv8n & 640 x 640 & 3.00 & 8.1\\
        YOLOv8s & 640 x 640 & 11.12 & 28.4\\
        YOLOv8m & 640 x 640 & 25.84 & 79.3\\
        YOLOv8l & 640 x 640 & 43.60 & 164.8\\
        \hline
    \end{tabular}
\end{table}

\begin{table}[tb!]
    \centering
    \caption{Comparing models' performance (inference time (ms), mAP@50 and mAP@50:95).}
    \label{tab:result2}
    
    \rowcolors{2}{gray!25}{white}
    \begin{tabular}{|c|c|c|c|} \hline 
        \rowcolor{gray!50}
        \textbf{Model} & \textbf{Inference Time (ms)} & \textbf{mAP@50} & \textbf{mAP@50:95} \\
        \hline
        YOLOv5n & 7.9 & 0.814 & 0.401\\
        YOLOv5s & 10.6 & 0.825 & 0.431\\
        YOLOv5m & 13.8 & 0.830 & 0.449\\
        YOLOv5l & 20.9 & 0.840 & 0.457\\
        YOLOv5x & 34.7 & 0.829 & 0.466\\
        YOLOv6n & 7.41 & 0.832 & 0.458\\
        YOLOv6s & 9.36 & 0.822 & 0.467\\
        \textbf{YOLOv6m} & \textbf{14.06} & \textbf{0.853} & \textbf{0.469}\\
        YOLOv6l & 18.06 & 0.854 & 0.473\\
        YOLOv6n6 & 15.81 & 0.816 & 0.457\\
        YOLOv6s6 & 25.06 & 0.828 & 0.449\\
        \textbf{YOLOv6m6} & \textbf{39.33} & \textbf{0.872} & \textbf{0.475}\\
        YOLOv6l6 & 57.46 & 0.848 & 0.474\\
        \textbf{YOLOv7tiny} & \textbf{7.5} & \textbf{0.837} & \textbf{0.433}\\
        YOLOv7 & 15.2 & 0.833 & 0.453\\
        YOLOv7x & 21.7 & 0.844 & 0.458\\
        YOLOv7w6 & 13.2 & 0.835 & 0.415\\
        YOLOv7e6 & 19.7 & 0.849 & 0.456\\
        YOLOv7d6 & 23.0 & 0.840 & 0.458\\
        \textbf{YOLOv8n} & \textbf{5.3} & \textbf{0.803} & \textbf{0.442}\\
        YOLOv8s & 8.2 & 0.802 & 0.442\\
        YOLOv8m & 11.2 & 0.817 & 0.464\\
        YOLOv8l & 14.5 & 0.790 & 0.438\\
        \hline
    \end{tabular}
\end{table}

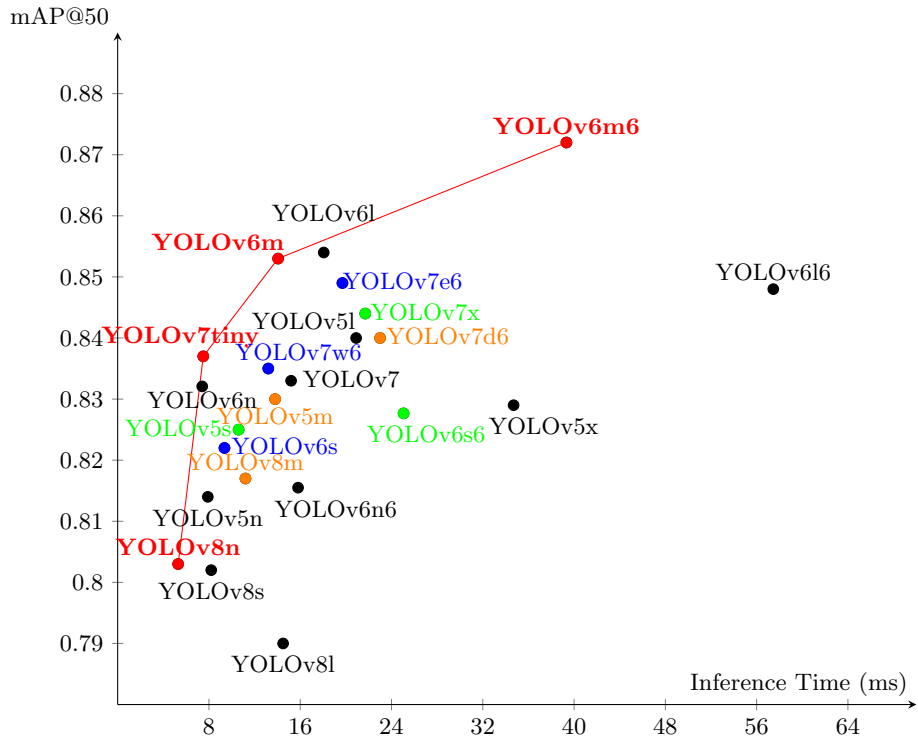
\begin{figure}
\centering
\begin{tikzpicture}[]
\begin{axis}[
  axis x line=center,
  axis y line=center,
  width={\linewidth},
  xtick={8,16,...,68},
  ytick={0.79, 0.80,...,0.88},
  xlabel={Inference Time (ms)},
  ylabel={mAP@50},
  xlabel style={},
  ylabel style={above left},
  xmin=0,
  xmax=70,
  ymin=0.78,
  ymax=0.89,
]

    \addplot[
    forget plot,
    mark=*,only marks,
    point meta =explicit symbolic,
    nodes near coords,
    visualization depends on={value \thisrow{shiftx} \as \myshiftx},
    visualization depends on={value \thisrow{shifty} \as \myshifty},
    visualization depends on={value \thisrow{color} \as \mycolor},
    every node near coord/.append style={yshift=\myshifty, xshift=\myshiftx},
    every node near coord/.append style={color=\mycolor},
    ] table[x=x,y=y,meta=label]{RR.dat};
    \addplot[mark=*, color=red] coordinates {(39.33,0.872) (14.06,0.853) (7.50,0.837) (5.30,0.803)};
    \addplot[only marks, mark=*, color=green] coordinates {(21.70, 0.84400) (25.06, 0.82766) (10.60, 0.82500)};
    \addplot[only marks, mark=*, color=blue] coordinates {(19.70, 0.84900) (13.20, 0.83500) (9.36, 0.82200)};
    \addplot[only marks, mark=*, color=orange] coordinates {(23.00, 0.84000) (13.80, 0.83000) (11.20, 0.81700)};
    
\end{axis}

\end{tikzpicture}
\caption{Precision-inference time curve of experimental models.}
\label{fig:map50}
\end{figure}

As shown in \autoref{tab:result2}, mAP@50:95 and mAP@50 fluctuated from 0.401 to 0.475 and from 0.790 to 0.872, respectively, while the inference time ranged from 5.3 ms to 57.46 ms. Based on mAP@50 and mAP@50:95, YOLOv6m6 obtained the highest accuracy, at $0.872$, $0.475$, while YOLOv8n observed the fastest inference time with $0.803$. According to~\autoref{fig:map50}, which shows the trade-off curve between mAP@50 and inference time, the four best models were chosen, including: YOLOv8n, YOLOv7tiny, YOLOv6m, YOLOv6m6.


\begin{figure}
\centering
\scalebox{0.8}{
\begin{tikzpicture}
    \begin{axis}[
        ybar,
        width  = \linewidth,
        height = 7cm,
        bar width=20pt,
        ylabel={Image/sec},
        symbolic x coords={YOLOv6m6, YOLOv6m, YOLOv7tiny, YOLOv8n},
        xtick = data,
        nodes near coords,
        enlarge y limits={value=0.2,upper},
        legend pos=north west,
    ]
    \addplot[fill=blue] coordinates {(YOLOv6m6, 1.246) (YOLOv6m, 5.512) (YOLOv7tiny, 36.308) (YOLOv8n, 58.272)};
    \end{axis}
\end{tikzpicture}
}
\caption{The processing rates (images per second) comparison of the four best trade-off models on Jetson Nano.}
\label{fig:table3}
\end{figure}
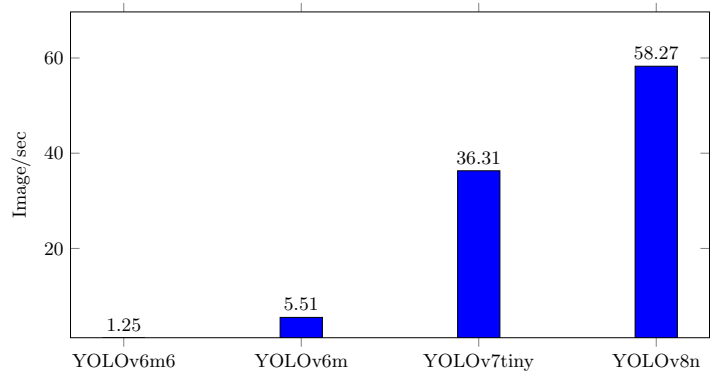

We then deployed these four best models on edge devices, represented by Jetson Nano, which can attached to UAVs for automatic visual inspection.~\autoref{fig:table3} illustrated that only YOLOv7tiny and YOLOv8n can achieve real-time processing rates of 36.308 and 58.272 images per second, respectively. Meanwhile, YOLOv6m6 and YOLOv6m have better mAP but reach limited processing rates at 1.246 and 5.512, in that order. 

\section{Conclusion}
Our research benchmarked the latest models in the YOLO family on the COCO-Bridge-2021+ dataset, partly assisting inspectors in choosing the most suitable model for the detail detection problem. The results point out that YOLOv5n, YOLOv7tiny, YOLOv6m, and YOLOv6m6 are the best models with high accuracy and inference speed. Our work paves the way for a new field: bridge details detection. There is considerable potential for refining and advancing our work. The number of benchmarked models is still limited, and the models' architectures are not optimized for the problem. Future works might explore the potential of various new architectures and modifications on known architectures. A performance improvement can also be made through better preprocessing and feature engineering using domain knowledge. More data can also lead to better real-life performance as it helps improve models' generalization and reduce overfitting. Better bridge detail detection helps contextualize the environment and defects, leading to better visual inspection of bridges.

\section*{Acknowledgment}
This research is funded by the University of Information Technology-Vietnam National University HoChiMinh City under grant number D1-2024-08.

\bibliography{conference_041818} 
\bibliographystyle{ieeetr}

\end{document}